%% file: acl_latex.tex
\pdfoutput=1

\documentclass[11pt]{article}

\usepackage[final]{acl}

\usepackage{times}
\usepackage{latexsym}

\usepackage[T1]{fontenc}

\usepackage[utf8]{inputenc}

\usepackage{microtype}

\usepackage{inconsolata}

\usepackage{graphicx}
\usepackage{booktabs}       
\usepackage[labelfont=bf,format=plain,justification=raggedright,singlelinecheck=false]{caption}
\usepackage{subcaption}

\usepackage[fleqn]{amsmath}
\usepackage{algorithm}
\usepackage{algpseudocode}
\usepackage{makecell}
\usepackage{anyfontsize}

\title{Towards LLMs Robustness to Changes in Prompt Format Styles}

\author{
 \textbf{Lilian Ngweta\textsuperscript{1}},
 \textbf{Kiran Kate\textsuperscript{2}},
 \textbf{Jason Tsay\textsuperscript{2}},
 \textbf{Yara Rizk\textsuperscript{2}}
\\
\\
 \textsuperscript{1}Rensselaer Polytechnic Institute,
 \textsuperscript{2}IBM Research
\\
 \small{
   \textbf{Correspondence:} \href{mailto:ngwetl@rpi.edu}
   {ngwetl@rpi.edu}
 }
}

\begin{document}
\maketitle
\begin{abstract}
Large language models (LLMs) have gained popularity in recent years for their utility in various applications. However, they are sensitive to non-semantic changes in prompt formats, where small changes in the prompt format can lead to significant performance fluctuations. In the literature, this problem is commonly referred to as prompt brittleness. Previous research on prompt engineering has focused mainly on developing techniques for identifying the optimal prompt for specific tasks. Some studies have also explored the issue of prompt brittleness and proposed methods to quantify performance variations; however, no simple solution has been found to address this challenge. We propose Mixture of Formats (MOF), a simple and efficient technique for addressing prompt brittleness in LLMs by diversifying the styles used in the prompt few-shot examples. MOF was inspired by computer vision techniques that utilize diverse style datasets to prevent models from associating specific styles with the target variable. Empirical results show that our proposed technique reduces style-induced prompt brittleness in various LLMs while also enhancing overall performance across prompt variations and different datasets. 
\end{abstract}


\input{ACL2025/sections/introduction}

\input{ACL2025/sections/related_work}

\input{ACL2025/sections/mof}

\input{ACL2025/sections/conclusion}



\bibliography{custom}

\newpage
\onecolumn
\appendix

\input{ACL2025/sections/appendix}

\end{document}

%% file: ACL2025/sections/introduction.tex
\section{Introduction}
\label{intro}

Large language models (LLMs) are useful for many applications and tasks i.e., content generation, translation, text analysis, etc. One of the popular techniques for adapting pre-trained LLMs to specific tasks that has emerged in recent years is prompt engineering \citep{liu2023pre, tonmoy2024comprehensive, chen2023unleashing}. Prompt engineering involves carefully crafting task-specific instructions and a few input-output demonstrations (prompts) to guide LLMs without changing their parameters \citep{sahoo2024systematic}. The popularity of prompt engineering can be attributed to the fact that it does not require labeled data and only needs a few demonstrations in prompts containing few-shot examples \citep{liu2023pre}. Prompting is also generally computationally cheaper than supervised fine-tuning techniques since the model parameters are not modified \citep{sahoo2024systematic}. 

Existing prompting techniques include zero-shot prompting \citep{radford2019language}, few-shot prompting \citep{brownfewshotlearners}, chain-of-thought (CoT) prompting \citep{wei2022chain}, and automatic chain-of-thought (Auto-CoT) prompting \citep{zhang2023automatic}. Most research on prompting techniques has focused on identifying or designing good prompts for specific tasks \citep{zhou2023large, wan-etal-2023-better}. However, a key problem often overlooked by these techniques is the sensitivity of LLMs to meaning-preserving changes in prompts. Examples of such changes include adding extra spaces, replacing two colons with one, changing the order of few-shot examples, or varying the choice of few-shot examples \citep{he2024does, sclar2023quantifying, lu-etal-2022-fantastically, wan-etal-2023-better}. This problem is sometimes referred to as prompt brittleness \citep{zhou2023batch}. Prompt brittleness contributes to LLMs being unreliable and prevents their adoption in high-risk domains such as healthcare.  

\begin{figure}[h]
    \centering
    \includegraphics[scale=0.22]{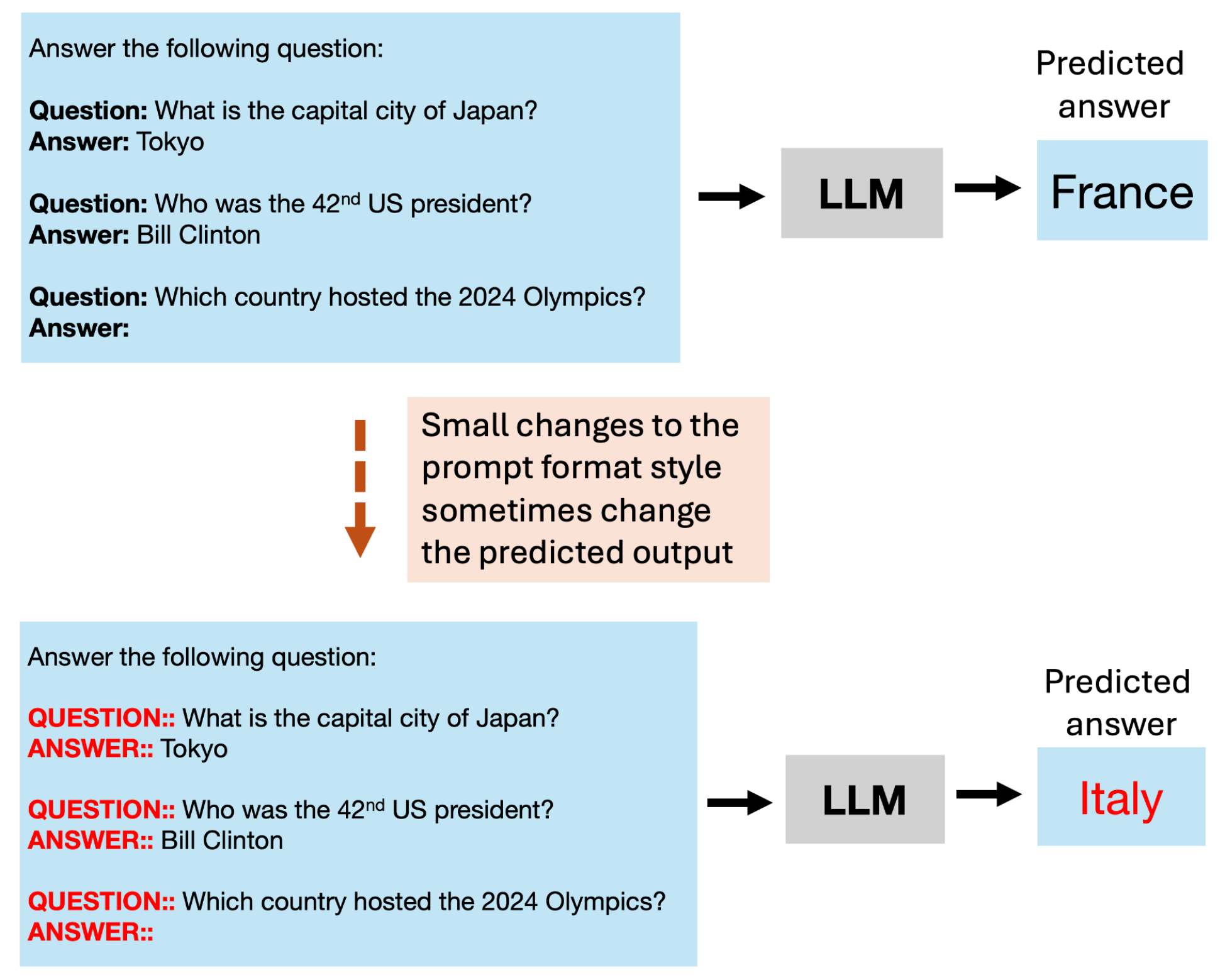}
    \caption{A demonstration of how small changes to the prompt format style can sometimes lead to incorrect predictions in LLMs.}
    \label{fig:problem}
\end{figure}

In this work, we focus on style-induced prompt brittleness as illustrated in Figure \ref{fig:problem}, and propose \textit{Mixture of Formats (MOF)} to address it. MOF is a simple and computationally efficient prompting technique where each few-shot example in the prompt is presented in a distinct style. Furthermore, the model is instructed to rewrite each example using a different style, as shown in Figure \ref{fig:mof}. MOF was inspired by ideas from computer vision that involve learning from datasets with diverse styles to prevent models from associating styles with the target variable \citep{arjovsky2019invariant, kamath2021does, yin2021optimization, wald2021calibration, ngweta2023simple, li2021searching}. We evaluate the effectiveness of MOF prompting using datasets from various tasks within SuperNaturalInstructions \citep{supernaturalinstructions}, comparing its performance against \textit{traditional prompts}. Our experiments focus on few-shot prompting, where a \textit{traditional prompt} refers to a regular few-shot prompt, and a \textit{MOF prompt} is a few-shot prompt that has been converted into the MOF style, as demonstrated in Figure \ref{fig:mof}.

%% file: ACL2025/sections/related_work.tex
\section{Related work}
\label{related_work}

\begin{figure*}[t]
    \centering
    \includegraphics[scale=0.32]{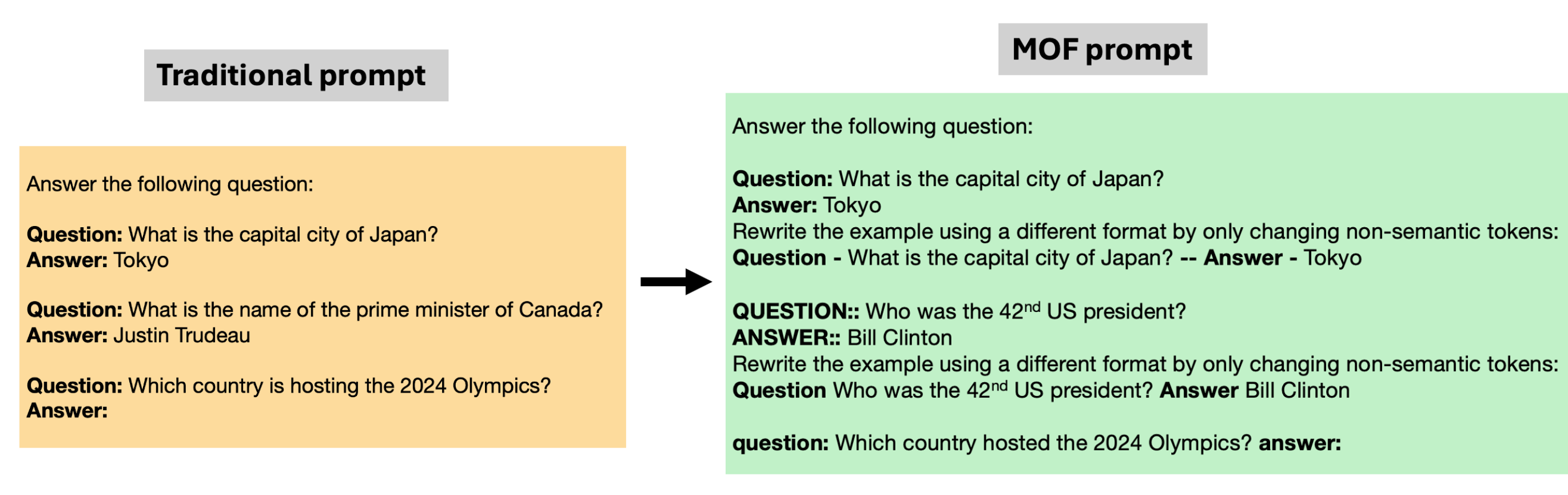}
    \caption{An illustration of how to convert a traditional prompt into a MOF prompt. This example serves as a simple demonstration of the conversion process. In the actual experiments, datasets use various formats such as \texttt{Passage:: \{\} , Answer:: \{\}} for dataset \textbf{task280}, \texttt{SYSTEM REFERENCE : \{\}. ORIGINAL
REFERENCE : \{\}. ANSWER : \{\}} for dataset \textbf{task1186}, and \texttt{Tweet:\{\} , Label:\{\} , Answer:\{\}} for dataset \textbf{task905}. These formats are generated using FormatSpread \citep{sclar2023quantifying}, as described in Section \ref{exp}. The datasets used are described in Table \ref{tab:datasets}.}
    \label{fig:mof}
\end{figure*}


\begin{figure*}
\captionsetup[subfigure]{justification=centering}
     \centering
     \begin{subfigure}[b]{0.48\textwidth}
         \centering
         \includegraphics[scale=0.43]{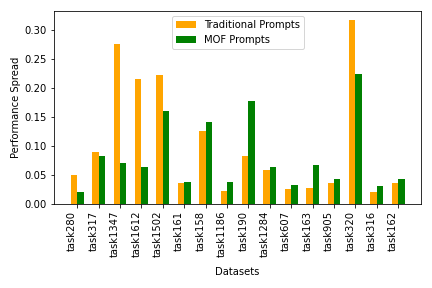}
         \caption{llama-3-70b-instruct}
         \label{fig:llama_3_70b_instruct}
     \end{subfigure}
     \hfill
     \begin{subfigure}[b]{0.48\textwidth}
         \centering
         \includegraphics[scale=0.43]{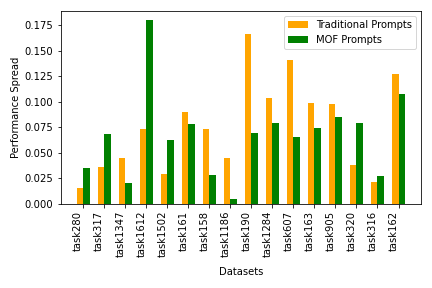}
         \caption{llama-2-13b-chat}
         \label{fig:llama_2_13b_chat}
     \end{subfigure}
     \hfill
     \begin{subfigure}[b]{0.48\textwidth}
         \centering
         \includegraphics[scale=0.43]{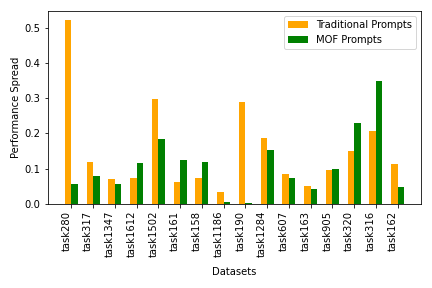}
         \caption{llama-2-13b}
         \label{fig:llama_2_13b}
     \end{subfigure}
     \hfill
     \begin{subfigure}[b]{0.48\textwidth}
         \centering
         \includegraphics[scale=0.43]{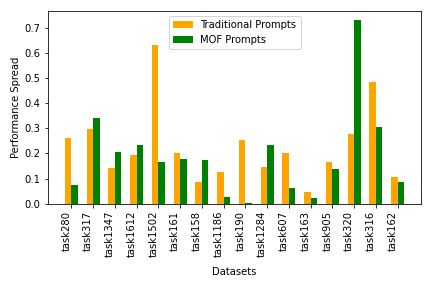}
         \caption{falcon-11B}
         \label{fig:falcon_11B}
     \end{subfigure}
         \centering
         \caption{Comparing the performance \textit{spread} of traditional prompts and MOF prompts. \textit{Spread} is a metric for quantifying style-induced prompt brittleness and it is obtained by taking the difference between the best performing prompt (maximum accuracy) and the worst performing prompt (minimum accuracy). MOF prompts perform comparably or outperform traditional prompts in most datasets and in some datasets, traditional prompts have better performance.}
         \label{fig:spread_bar_plots}
        
\end{figure*}


\paragraph{Traditional prompt engineering techniques.} Several prompt engineering techniques have been proposed in recent years. Zero-shot prompting is a technique in which a prompt contains a description of the task and no training data is required \citep{radford2019language}. Unlike zero-shot prompting, few-shot prompting adds a few input-output demonstrations to the prompt to further help the model understand the task \citep{brownfewshotlearners}. Both zero-shot and few-shot prompting techniques enable the application of LLMs on new tasks without extensive training \citep{sahoo2024systematic}. For reasoning and logic tasks, prompting techniques that have been proposed include chain-of-thought (CoT) \citep{wei2022chain} and automatic chain-of-thought (Auto-CoT) \citep{zhang2023automatic}. CoT is a prompting technique that encourages LLMs to do step-by-step reasoning \citep{wei2022chain}. Since manually creating CoT examples is time-consuming and not easily scalable, \citet{zhang2023automatic} proposed Auto-CoT to automatically guide LLMs to generate reasoning steps using a "Let's think step by step" statement in the prompt. 

These traditional prompting techniques can be adapted to the MOF format by applying different formatting styles to each prompt example, as demonstrated in Figure \ref{fig:mof}. In this paper, we focus on the application of MOF to few-shot prompting.

\paragraph{Optimizing for the best prompt.} This line of work focuses on optimizing and identifying the most effective prompt for a given task. \citet{zhou2023large} propose the automatic prompt engineer (APE), an approach that enables the generation and selection of prompt instructions automatically. APE involves analyzing input queries, generating candidate prompt instructions, and then using reinforcement learning to select the best prompt \citep{zhou2023large}.  Similarly, \citet{wan-etal-2023-better} propose a method where an LLM generates zero-shot outputs for given inputs, followed by selecting high-quality few-shot examples to construct an improved prompt, focusing on consistency, diversity, and repetition. Since automatic prompt optimization (APO) methods focus on optimizing instruction or optimizing few-shot examples, \citet{wan2024teach} propose a technique to optimize for both, and compare its performance with the performance of techniques that only optimize instructions or examples. \citet{yang2024largelanguagemodelsoptimizers} present Optimization by PROmpting (OPRO), a method that leverages LLMs as optimizers by describing the optimization task in natural language \citep{yang2024largelanguagemodelsoptimizers}. \citet{pryzant2023automatic} propose Prompt Optimization with Textual Gradients (ProTeGi), which employs text gradients guided by beam search and bandit selection techniques for automatic prompt optimization \citep{pryzant2023automatic}. Additionally, \citet{khattab2024dspy} introduce DSPy, a framework that replaces hard-coded prompt templates with a systematic approach for building language model pipelines. Other methods for identifying optimal prompts include \citep{feffer2024prompt, sorensen-etal-2022-information, yin-etal-2023-read}. 

Unlike existing methods in this area that repeatedly search for optimal prompts per task and model, our goal is to reduce style-induced prompt brittleness using an efficient and straightforward recipe illustrated in Figure \ref{fig:mof}.  

\paragraph{Quantifying prompt brittleness in LLMs.} Several works have shown that LLMs are sensitive to changes in prompt formats \citep{sclar2023quantifying, he2024does, voronov2024mind} and to the order of few-shot examples in the prompt \citep{lu-etal-2022-fantastically}. \citet{sclar2023quantifying} propose FormatSpread, a method to efficiently measure performance variations in LLMs caused by prompt format changes, by computing the performance difference (\textit{spread}) between the best-performing format and the worst-performing format. Due to the sensitivity of LLMs to prompt format variations, \citet{polo2024efficient} propose PromptEval, an efficient method for evaluating LLMs on multiple prompts instead of a single prompt. Similarly, \citet{mizrahi2024state} propose metrics for multi-prompt evaluation of LLMs.

While these approaches are valuable tools for quantifying prompt brittleness, our proposed method focuses on mitigating it, particularly the brittleness arising from style variations in prompt formats.

\paragraph{Prompt ensembles.} \citet{arora2022ask} introduce Ask Me Anything (AMA), a prompting approach that transforms inputs into a question-answering format to encourage open-ended responses. AMA generates multiple imperfect prompts and combines the responses using a weak supervision strategy to produce the final output \citep{arora2022ask}. Similarly, \citet{voronov2024mind} propose Template Ensembles, an approach that aggregates model predictions across multiple prompt templates. However, both methods are computationally expensive, as they require aggregating predictions from multiple prompts. Furthermore, unlike our proposed method, they do not specifically address prompt brittleness caused by style variations in prompt formats.

%% file: ACL2025/sections/mof.tex
\section{Mixture of Formats}
\label{mof}

Style-induced prompt brittleness in LLMs is similar to problems observed in computer vision, where small changes to an image's style (eg. color or background) can affect the model's ability to make accurate predictions \citep{nagarajan2020understanding}. In computer vision, various approaches have been developed to address this issue, often involving learning from diverse datasets \citep{arjovsky2019invariant, ngweta2023simple, kamath2021does, yin2021optimization, wald2021calibration, li2021searching}. The underlying idea is that exposure to diverse data points helps the model disassociate styles from the target variable. Drawing inspiration from these techniques, we propose Mixture of Formats (MOF), a novel prompting strategy that deviates from traditional ways of crafting prompts by employing a distinct style format for each few-shot example in the prompt. To further reinforce model understanding, we have the model rewrite the question and answer of each example using a different format style, as illustrated in Figure \ref{fig:mof}. The effectiveness of this approach is evaluated in the subsequent subsections.

\subsection{Experiments}
\label{exp}


\begin{table*}[h]
    \centering
    \caption{Best performing format (\textit{Max Accuracy}) and worst performing format (\textit{Min Accuracy}) results for both traditional prompts and MOF prompts for \texttt{llama-3-70b-instruct}. MOF prompts improve the \textit{Min Accuracy} and the \textit{Max Accuracy} over traditional prompts in most cases.}
    \label{tab:min_max_acc}
    \begin{tabular}{ccc|ccc}
    \toprule
           Task & \multicolumn{2}{c}{Traditional Prompts} & \multicolumn{2}{c}{MOF Prompts} \\
           {} &   Min Accuracy & Max Accuracy & Min Accuracy & Max Accuracy \\
    \midrule
       task280 &               0.811 &        0.860 &        \textbf{0.880} &        \textbf{0.900} \\
       task317 &               0.139 &        0.229 &        \textbf{0.712} &        \textbf{0.795} \\
      task1347 &               0.248 &        0.524 &        \textbf{0.464} &        \textbf{0.535} \\
      task1612 &               0.624 &        0.839 &        \textbf{0.787} &        \textbf{0.851} \\
      task1502 &               0.443 &        \textbf{0.666} &        \textbf{0.479} &        0.639 \\
       task161 &               0.472 &        0.507 &        \textbf{0.475} &        \textbf{0.512} \\
    \bottomrule
    \end{tabular}
\end{table*}


Let $X$ denote input queries for a task, and $Y$ denote the target variable. Given $N$ observations of inputs $X$ and their corresponding targets $Y$ as data $\mathcal{D} = \{X_n, Y_n\}_{n=1}^N$, we automatically build a traditional prompt and its MOF prompt version, each containing 5 few-shot examples, and use them for inference with an LLM. The traditional prompt is created using FormatSpread \citep{sclar2023quantifying}, while the MOF prompt is generated by modifying FormatSpread to incorporate diverse formats within the few-shot examples, as illustrated in Figure \ref{fig:mof}. 

Using FormatSpread, we create 10 traditional prompt variations and 10 MOF prompt variations. From the 10 prompt variations, for both traditional and MOF prompts, we compute performance accuracies for each prompt format across various tasks. The goal is to compare the style-induced prompt brittleness between traditional prompts and MOF prompts. As in \citet{sclar2023quantifying}, we measure brittleness by calculating the performance \textit{spread}, defined as the accuracy difference between the best-performing and worst-performing prompt formats. The evaluation pipelines for traditional and MOF prompts are summarized in Algorithm \ref{algo:trad_eval_algorithm} and Algorithm \ref{algo:mof_eval_algorithm}, respectively.

\paragraph{Datasets} We perform experiments on datasets covering various tasks from SuperNaturalInstructions \citep{naturalinstructions, supernaturalinstructions}. Due to limited computational resources, we randomly selected 16 datasets and for each dataset we use 1000 samples and a batch size of 100. The datasets used are described in Table \ref{tab:datasets}.

\paragraph{Baselines, metrics, and LLMs used}
In our experiments, we use traditional few-shot prompts as our baselines, where we compare the performance of LLMs when using traditional prompts versus MOF prompts. A primary focus of this work is to determine whether MOF prompting can minimize performance variations (\textit{spread}) in LLMs when prompt format styles change. The performance \textit{spread} is obtained by taking the difference between the highest performing prompt (denoted as "Max Accuracy" in the results tables) and the minimum performing prompt (denoted as "Min Accuracy"). The \textit{spread} value ranges from 0.0 to 1.0, where values closer to 0.0 indicate that the LLM is more robust and less sensitive to style changes, while values closer to 1.0 suggest that the LLM is highly sensitive to these changes. Additionally, for both traditional and MOF prompts, we compute the average accuracy across all 10 prompt variations to assess the overall performance of MOF prompts relative to traditional prompts. We use four LLMs in our experiments: \texttt{falcon-11B}, \texttt{Llama-2-13b-hf}, \texttt{Llama-2-13b-chat-hf}, and \texttt{llama-3-70b-instruct}. 

We emphasize that while MOF prompting can be applied and compared with other existing traditional prompting techniques, such as automatic chain-of-thought (Auto-CoT) \citep{zhang2023automatic} and the automatic prompt engineer (APE) \citep{zhou2023large}, this paper focuses on applying MOF prompting to regular few-shot prompting and comparing their performances, due to limited computational resources. 

\paragraph{Generating responses for evaluation} To generate a response for a given question, a traditional or MOF prompt is combined with the question and then passed to an LLM to generate the response. The generated response is then compared to the ground-truth answer to calculate the model's accuracy.

\subsection{Results}
We perform experiments to evaluate whether MOF prompts reduce prompt brittleness in LLMs by comparing their \textit{spread} with traditional prompts. We also assess improvements by analyzing the best (Max Accuracy) and worst (Min Accuracy) performing prompts. Finally, we evaluate overall performance by comparing the mean accuracies across all 10 prompt variations for both prompt types.  

\paragraph{Minimizing prompt brittleness} Figure \ref{fig:spread_bar_plots} shows that MOF prompting effectively reduces style-induced prompt brittleness across several datasets and LLMs, with a notable 46\% reduction in \texttt{task280} using \texttt{Llama-2-13b}. While MOF prompts generally perform as well or better than traditional prompts, exceptions occur in \texttt{task190} (\texttt{llama-3-70b-instruct}), \texttt{task1612} (\texttt{llama-2-13b-chat}), and \texttt{task320} (\texttt{falcon-11B}), where traditional prompts perform better. Investigating why MOF fails on these datasets is an important future direction.


\paragraph{Best and worst performing prompts} Results for the best-performing prompt (Max Accuracy) and worst-performing prompt (Min Accuracy) for both traditional and MOF prompting are reported in Table \ref{tab:min_max_acc}. We observe that MOF prompting not only reduces spread but also improves both minimum and maximum accuracies. Average accuracy results across all 10 prompt variations for both traditional and MOF prompts are discussed in Appendix \ref{app}.

%% file: ACL2025/sections/conclusion.tex
\section{Conclusion and future work}
\label{conclusion}
Addressing prompt brittleness remains a challenge, particularly when caused by changes in prompt format styles. In this work, we introduce a simple and efficient prompting technique, MOF, and evaluate its effectiveness in addressing style-induced prompt brittleness. The preliminary results are promising, with significant improvements over traditional prompting in many datasets, as shown in Figure \ref{fig:spread_bar_plots}. 

Future directions include integrating MOF with techniques like chain-of-thought (CoT) and automatic prompt engineer (APE), comparing its performance with methods that aggregate results from multiple prompts such as AMA \citep{arora2022ask} and Template Ensembles \citep{voronov2024mind}, and conducting experiments with larger LLMs like GPT-4, Claude 3.5 Sonnet, Falcon 40B, and Llama 3.1 405B. Additionally, analyzing MOF's failures on certain datasets is a crucial area for further exploration.


We hope this work will inspire further research into addressing prompt brittleness in LLMs, and the code for this project is publicly available on GitHub.\footnote{Code: \href {https://github.com/lilianngweta/mof}{github.com/lilianngweta/mof}.}


%% file: ACL2025/sections/appendix.tex
\section{Appendix}
\label{app}

\paragraph{Average accuracy across all 10 prompt variations} Up to this point, we have examined the performance in minimizing prompt brittleness, as well as the performance of the best and worst performing prompts. In this section, we focus on the performance of traditional and MOF prompts across all 10 prompt variations for each. The average accuracy across these 10 prompt variations for both traditional and MOF prompts is reported in Table \ref{tab:avg_accuracies}. For all LLMs, we find that MOF prompts perform nearly as well as traditional prompts, with MOF prompts generally leading to significant overall mean accuracy improvements. \\

\begin{algorithm}[H]
    \centering
    \caption{Traditional prompts evaluation pipeline}
    \label{algo:trad_eval_algorithm}
    \begin{algorithmic}[1] 
        \State \textbf{Input}: Data $\mathcal{D}$
        \State Create 10 variations of traditional prompts using FormatSpread \citep{sclar2023quantifying}. 
        \State Use the created traditional prompt variations to generate responses.
        \State Evaluate each of the 10 traditional prompts and save results.
        \State Compute the average accuracy across all 10 traditional prompt variations.
        \State Identify the best performing prompt, the worst performing prompt, and compute the spread. 
        \State \textbf{Output}: Return accuracies for the best performing prompt (max accuracy), worst performing prompt (min accuracy), the spread, and the average accuracy across all 10 traditional prompt variations.
        
    \end{algorithmic}
\end{algorithm}

\begin{algorithm}[H]
    \centering
    \caption{MOF prompts evaluation pipeline}
    \label{algo:mof_eval_algorithm}
    \begin{algorithmic}[1] 
        \State \textbf{Input}: Data $\mathcal{D}$
        \State Create 10 variations of MOF prompts using a \textbf{modified} FormatSpread \citep{sclar2023quantifying} that incorporates diverse styles in the few-shot examples as illustrated in Figure \ref{fig:mof}.
        \State Use the created MOF prompt variations to generate responses.
        \State Evaluate each of the 10 MOF prompts and save results.
        \State Compute the average accuracy across all 10 MOF prompt variations.
        \State Identify the best performing prompt, worst performing prompt, and compute the spread.
        \State \textbf{Output}: Return accuracies for the best performing prompt (max accuracy), worst performing prompt (min accuracy), the spread, and the average accuracy across all 10 MOF prompt variations.
        
    \end{algorithmic}
\end{algorithm}



\begin{table*}[h]
    \caption{Average accuracy results across 10 prompt variations for traditional prompts (denoted as \textit{Trad Mean Acc}) and MOF prompts (denoted as \textit{MOF Mean Acc}). For all LLMs, MOF prompts perform comparable and in most cases have a higher overall average accuracy than traditional prompts.}
     \label{tab:avg_accuracies}
    \begin{subtable}[h]{0.48\textwidth}
        \centering
        \caption{Llama-2-13b-chat}
       \label{tab:llama-2-13b-chat-hf}
        \begin{tabular}{lccc}
        \toprule
               Task &  Trad Mean Acc &  MOF Mean Acc \\
        \midrule
           task280 &           \textbf{0.853} &                      0.841 \\
           task317 &           0.578 &                      \textbf{0.749} \\
          task1612 &           0.471 &                      \textbf{0.490} \\
          task1502 &           \textbf{0.596} &                      0.579 \\
          task161 &           0.199 &                      \textbf{0.278} \\
        \bottomrule
        \end{tabular}
    \end{subtable}
    \hfill
    \begin{subtable}[h]{0.48\textwidth}
        \centering
        \caption{Llama-2-13b}
        \label{tab:llama-2-13b-hf}
        \begin{tabular}{lccc}
        \toprule
               Task &  Trad Mean Acc &  MOF Mean Acc \\
        \midrule
           task280 &           0.635 &                      \textbf{0.842} \\
           task317 &           0.564 &                      \textbf{0.725} \\
          task1612 &           \textbf{0.564} &                      0.505 \\
          task1502 &           \textbf{0.489} &                      0.485 \\
           task161 &           0.245 &                      \textbf{0.371} \\
        \bottomrule
        \end{tabular}
     \end{subtable}
     
     \noindent
    \begin{subtable}[h]{0.48\textwidth}
        \centering
        \vspace{2em}
        \caption{falcon-11B}
        \label{tab:falcon-11B}
        \begin{tabular}{lccc}
        \toprule
               task &  Trad Mean acc &  MOF Mean acc \\
        \midrule
           task280 &           0.727 &                      \textbf{0.802} \\
           task317 &           0.501 &                      \textbf{0.672} \\
          task1612 &           \textbf{0.638} &                      0.553 \\
          task1502 &           0.305 &                      \textbf{0.493} \\
           task161 &           \textbf{0.390} &                      0.387 \\
        \bottomrule
        \end{tabular}
     \end{subtable}
     \hfill
    \begin{subtable}[h]{0.48\textwidth}
        \centering
        \vspace{2em}
        \caption{llama-3-70b-instruct}
        \label{tab:llama-3-70b-instruct}
        \begin{tabular}{lccc}
        \toprule
               task &  Trad Mean acc &  MOF Mean acc \\
        \midrule
           task280 &           0.836 &                      \textbf{0.890} \\
           task317 &           0.154 &                      \textbf{0.770} \\
          task1612 &           0.800 &                      \textbf{0.821} \\
          task1502 &           \textbf{0.600} &                      0.593 \\
           task161 &           \textbf{0.496} &                      0.492 \\
        \bottomrule
        \end{tabular}
     \end{subtable}
     
\end{table*}


\begin{table*}[h]
    \centering
    \scriptsize
\caption{Datasets from SuperNaturalInstructions \citep{naturalinstructions, supernaturalinstructions} that we used in our experiments.}
\label{tab:datasets}
    \renewcommand{\arraystretch}{1.7}
    \begin{tabular}{|l|l|}
        \hline
        \textbf{Dataset ID} & \textbf{Dataset Description}\\ \hline
        task280& \makecell[l]{A text categorization dataset that involves classifying sentences into four types of stereotypes: gender, profession, \\race, and religion.}\\ \hline
        task317 & \makecell[l]{A stereotype detection dataset that involves classifying sentences into various types of stereotypes.} \\ \hline
        task1347 & \makecell[l]{A text matching dataset that involves classifying the semantic similarity of two sentences on a scale of 0 - 5.} \\ \hline
        task1612 & \makecell[l]{A textual entailment dataset derived from the SICK dataset, that involves accurately classifying labels to show the \\relationship between two sentences.} \\ \hline
        task1502 & \makecell[l]{A toxic language detection dataset that involves classifying the type of tweet in HateXplain.} \\ \hline
        task161 & \makecell[l]{A dataset focused on counting the words in a sentence that contain a specified letter.} \\ \hline
        task158 & \makecell[l]{A dataset that involves counting the number of times a word occurs in a sentence.} \\ \hline
        task1186 & \makecell[l]{A text quality evaluation dataset that involves evaluating the naturalness of system generated reference.} \\ \hline
        task190 & \makecell[l]{A textual entailment dataset that involves choosing whether two given sentences agree, disagree, or neither with each other.} \\ \hline
        task1284 & \makecell[l]{A text quality evaluation dataset that involves evaluating the informativeness of system generated reference.} \\ \hline
        task607 & \makecell[l]{A toxic language detection that involves determining whether or not the post is intentionally offensive.} \\ \hline
        task163 & \makecell[l]{A dataset that involves counting the number of words in the sentence that end with a specified letter.} \\ \hline
        task905 & \makecell[l]{A toxic language detection dataset that involves determining whether the given category of a tweet is true or false.} \\ \hline
        task320 & \makecell[l]{A stereotype detection dataset that involves determining whether a given target pertaining to race in two sentences is \\a stereotype.} \\ \hline
        task316 & \makecell[l]{A stereotype detection dataset that involves classifying whether a sentence is stereotype or anti-stereotype.} \\ \hline
        task162& \makecell[l]{A dataset that involves counting the words in a sentence that begin with a specified letter.}\\ \hline
    \end{tabular}

\end{table*}